# A Visual Kinematics Calibration Method for Manipulator Based on Nonlinear Optimization☆

Gang Peng[a,b], Zhihao Wang[a,b], Jin Yang [a,b,]*, Xinde Li[c]

[a] Key Laboratory of Image Processing and Intelligent Control, Ministry of Education;
[b] School of Artificial Intelligence and Automation, Huazhong University of Science and Technology, Wuhan, China;
[c] IEEE senior member, School of Automation, South East University, Nanjing, China

**Abstract:** The traditional kinematic calibration method for manipulators requires precise three-dimensional measuring instruments to measure the end pose, which is not only expensive due to the high cost of the measuring instruments but also not applicable to all manipulators. Another calibration method uses a camera, but the system error caused by the camera's parameters affects the calibration accuracy of the kinematics of the robot arm. Therefore, this paper proposes a method for calibrating the geometric parameters of a kinematic model of a manipulator based on monocular vision. Firstly, the classic Denavit–Hartenberg (D-H) modeling method is used to establish the kinematic parameters of the manipulator. Secondly, nonlinear optimization and parameter compensation are performed. The three-dimensional positions of the feature points of the calibration plate under each manipulator attitude corresponding to the actual kinematic model and the classic D-H kinematic model are mapped into the pixel coordinate system, and the sum of Euclidean distance errors of the pixel coordinates of the two is used as the objective function to be optimized. The experimental results show that the pixel deviation of the end pose corresponding to the optimized D-H kinematic model proposed in this paper and the end pose corresponding to the actual kinematic model in the pixel coordinate system is 0.99 pixels. Compared with the 7.9 deviation pixels between the pixel coordinates calculated by the classic D-H kinematic model and the actual pixel coordinates, the deviation is reduced by nearly 7 pixels for an 87% reduction in error. Therefore, the proposed method can effectively avoid system errors caused by camera parameters in visual calibration, can improve the absolute positioning accuracy of the end of the robotic arm, and has good economy and universality.

## 1. Introduction

The calibration of the geometric parameters of the kinematic model of a manipulator is mainly divided into two categories: parameter calibration based on kinematic model (referred to as kinematic model calibration) and self-calibration. Parameter calibration based on the kinematic model is a method for obtaining geometric parameters of a more accurate kinematic model by parameter identification based on the established kinematic model and measurement data. Generally, a high-precision three-dimensional measuring device (such as a laser tracker) is required. The measuring device obtains the posture of the end of the robot arm, and the measured data has high precision and low noise, but since the device is expensive and requires a professional operation, it is not universal. Self-calibration can be divided into two categories: physical constraint calibration and additional

---

☆This work was supported by National Natural Science Foundation of China(No.91748106) and Hubei Province Natural Science Foundation of China(No. 2019CFB526).

* Corresponding author.

Peng Gang, PhD, Assoc. Prof, Email: penggang@hust.edu.cn; Wang Zhihao(Co-First Author), Master;

Yang Jin(Corresponding Author) Master graduate student, Email:m201972630@hust.edu.cn;

Li Xinde, PhD, Prof, IEEE senior member.

redundant sensor calibration. In physical constraint calibration, a number of closed-loop position equations are constructed by special end-effector contact with another plane (or point), and a model is used to solve for the kinematic parameters based on the established error. Additional redundant sensor calibration involves measuring the pose information of the end of the manipulator with a sensor fixed at the end of the manipulator and then solving the kinematic parameters of the manipulator [1].

In the early days of kinematic calibration technology development, some researchers [2] used circle point analysis (CPA) dot analysis to obtain the pose of each axis. With this approach, the geometric parameters of the actual kinematic model are determined by measuring geometrical parameters such as the spatial distance between the baselines of the joints and the angle of rotation. In the process of calibrating the kinematic model directly by the CPA method, since the position of the base coordinate system and the tool coordinate system of the robot arm cannot be directly obtained, the accuracy of the calibrated kinematic parameters is low, and the corrected end of the arm movement accuracy cannot meet the requirements [3]. Therefore, the commonly used robot calibration method mainly includes four steps: modeling, measurement and parameter identification.

(1) Modeling: The commonly used kinematic model is the Denavit–Hartenberg (D-H) model, which expresses the transformation relationship of two adjacent link coordinate systems by a homogeneous transformation matrix, but there are singular problems when two adjacent axes are parallel. Therefore, many scholars have proposed new kinematic modeling methods to increase universality. For example, Hayati [4] proposed the modified D-H model by correcting the micro-angular angles of the two parallel axes. Other solutions include the complete and parametrically continuous (CPC) kinematic model proposed by Zhang [5] and Schroer [6] and the zero reference model proposed by Azerounian [7].

(2) Measurement: The goal of measurement is to obtain the pose of the end of the manipulator through a high-precision three-dimensional measuring instrument. The measurement accuracy has an important influence on the calibration result, but the price of the high-precision measuring instrument is generally expensive, so a camera-based visual calibration system has also been proposed as a measurement scheme [8].

(3) Parameter identification: Parameter identification is the process of minimizing the error between the measured value of the end position of the arm and the actual value after calibration by the optimization algorithm. Zhang et al. [9] established the absolute positioning error equations related to the position of the robot's end and the spindle parameters and then made improvements. Some researchers have also developed innovations to improve the accuracy of parameter identification. Wang et al. [10] proposed a filtering-based multivariate innovative stochastic gradient algorithm that can achieve high-precision parameter identification. The computational complexity of current methods for parameter identification is high. This paper makes full use of the information obtained by the monocular vision system to reconstruct the optimization target, which improves the calibration accuracy of the kinematic parameters and reduces the computational complexity. Finally, the accurate geometric parameters after parameter identification are updated to the controller of the manipulator and control the motion of the robot arm as the actual value of the kinematic model.

The existing visual calibration method uses a camera for calibration, but system errors caused by camera parameters will affect the accuracy of the kinematic calibration of the robotic arm. Parameter identification is a key step in the calibration of the kinematic of the manipulator. Since there are many optimization variables to be identified, the selection of nonlinear optimization algorithms becomes the key to the calibration accuracy. Therefore, a robotic kinematic vision calibration method based on non-linear optimization is proposed. The information obtained by monocular vision is used to optimize the D-H kinematic parameters of the robotic arm. The three-dimensional positions of the feature points of the calibration plate in the attitude of each robotic arm corresponding to the actual kinematic model and the classic D-H kinematic model are mapped to the pixel coordinate system, and the Euclidean distance error between the two is used as the objective function to be optimized. Combined with rough identification based on axis measurement and nonlinear optimization, the precise identification method obtains the actual deviation of the kinematic parameters. Through parameter compensation, the optimized D-H kinematic model can effectively avoid system errors caused by camera parameters in visual calibration and improve the absolute positioning accuracy of the end of the robotic arm.

## 2. Method

### 2.1. Kinematic model

This paper uses the classic D-H method to establish a kinematic model [11]. The fixed coordinate system is fixed on the latter link, and a homogeneous transformation matrix $^{i-1}_{i}T$ established by four parameters ($\theta, d, a,$ and $\alpha$) describes the spatial geometric relationship between a certain coordinate system {i-1} and the previous link coordinate system {i}. That is,

$$^{i-1}_{i}T = \text{Rot}(z, \theta_i)\text{Trans}(z, d_i)\text{Trans}(x, a_i)\text{Rot}(x, \alpha_i) \tag{1}$$

The general expression of $^{i-1}_{i}T$ is

$$^{i-1}_{i}T = \begin{bmatrix} c\theta_i & -s\theta_i c\alpha_i & s\theta_i s\alpha_i & a_i c\theta_i \\ s\theta_i & c\theta_i c\alpha_i & -c\theta_i s\alpha_i & a_i s\theta_i \\ 0 & s\alpha_i & c\alpha_i & d_i \\ 0 & 0 & 0 & 1 \end{bmatrix} \tag{2}$$

where $s\theta_i = sin(\theta_i), c\theta_i = cos(\theta_i)$. Then, by multiplying the homogeneous transformation matrices $^{i-1}_{i}T (i = 1,2,...,n)$ of all the single links, the position from the base of the robot to the end position can be calculated. Homogeneous transformation matrix $^{0}_{n}T$:

$$^{o}_{n}T = {^{0}_{1}T}{^{1}_{2}T}...{^{n-1}_{n}T} \tag{3}$$

The D-H coordinate system is established according to the structure of the robot arm as shown in Fig. 1, and the corresponding D-H parameters are established as shown in Table 1. Therefore, it is possible to obtain $^{i-1}_{i}T$ of each link and a homogeneous transformation matrix $^{0}_{7}T$ of the base of the robot arm to the end position.

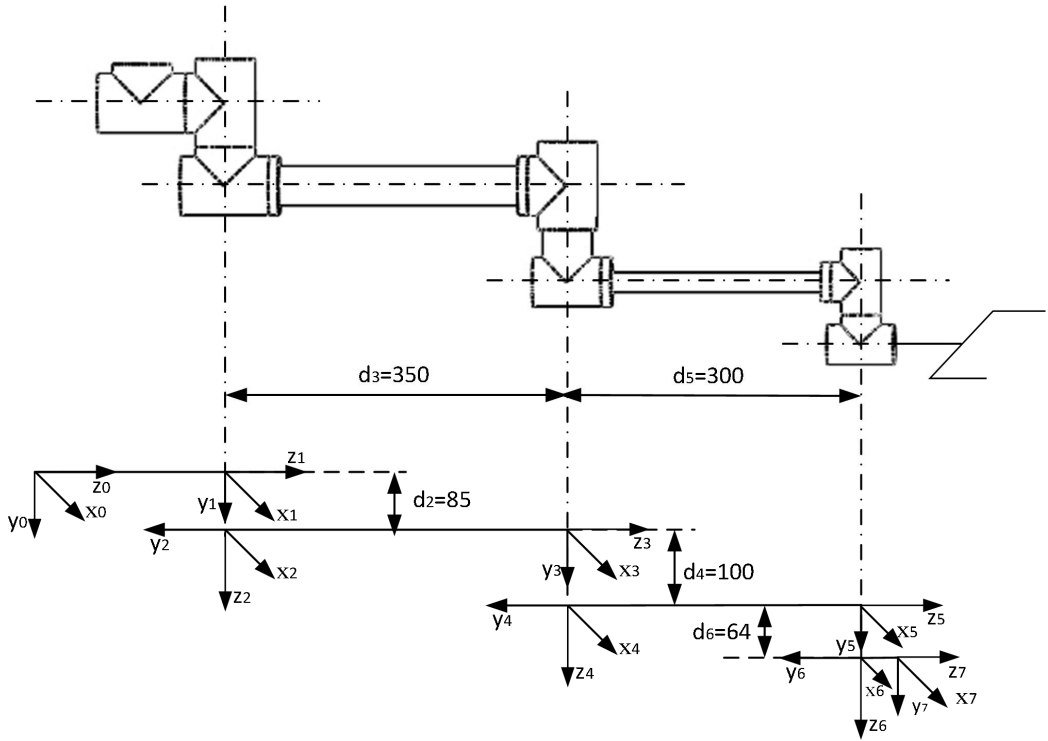

Fig. 1. Offset seven-degree-of-freedom mechanical arm linkage coordinate system.

Table 1. Offset seven-degree-of-freedom manipulator D-H parameter table.

| Joint $i$ | $\alpha_{i-1}(°)$ | $a_{i-1}$(mm) | $d_i$(mm) | $\theta_i(°)$ | Joint Range (°) |
|---|---|---|---|---|---|
| 1 | 0 | 0 | 85 | 0 | ± 180 |
| 2 | $-\pi/2$ | 0 | 85 | 0 | ± 180 |
| 3 | $\pi/2$ | 0 | 350 | 0 | ± 180 |
| 4 | $-\pi/2$ | 0 | 100 | 0 | ± 180 |
| 5 | $\pi/2$ | 0 | 300 | 0 | ± 180 |
| 6 | $-\pi/2$ | 0 | 64 | 0 | ± 180 |
| 7 | 0 | 0 | 42 | 0 | ±180 |

According to the D-H parameter table established in Table 1, there are a total of 4*7=28 parameters from the base coordinate system of the robot arm to the end of the seventh joint, and there is no coupling relationship between these parameters. The transformation from the measurement coordinate system to the base coordinate system of the robot arm $^B_C T$ can be expressed based on the translation amount $d_{b\_x}, d_{b\_y}, d_{b\_z}$ and the rotation amount $\delta_{b\_x}, \delta_{b\_y}, \delta_{b\_z}$ as shown in Eq. (4):

$$^B_C T = T_{xb}(d_{b\_x})T_{yb}(d_{b\_y})T_{zb}(d_{b\_z})R_{xb}(\delta_{b\_x})R_{yb}(\delta_{b\_y})R_{zb}(\delta_{b\_z}) \quad (4)$$

Similarly, the transformation $^T_W T$ from the end of the seventh joint (wrist coordinate system: {W}) to the tool coordinate system T can be expressed based on the translation amount $d_{t\_x}, d_{t\_y}, d_{t\_z}$ and the amount of rotation $\delta_{t\_x}, \delta_{t\_y}, \delta_{t\_z}$ as shown in Eq. (5):

$$^T_W T = T_{xt}(d_{t\_x})T_{yt}(d_{t\_y})T_{zt}(d_{t\_z})R_{xt}(\delta_{t\_x})R_{yt}(\delta_{t\_y})R_{zt}(\delta_{t\_z}) \quad (5)$$

The transformation from the end of the arm to the calibration plate $^T_W T$ is also coupled with the D-H parameter of the seventh joint, i.e., the translation amount $d_{t\_z}$ along the z-axis in $^T_W T$ and $d_7$ of the seventh joint. The parameter correlation and the translation amount $d_{t\_z}$ can be equivalent using the parameter $d_7$. Assume that the deviation of the rotation of the end of the calibration plate and the seventh joint of the robot arm is neglected, so the transformation relationship $^T_W T$ of the end of the robot arm to the calibration plate can be determined by only two parameters [12]. Therefore, the transformation relationship from the measurement coordinate system to the measurement point of the calibration plate can be determined by 28+6+2=36 independent parameters, and the transformation matrix $^T_C T$ from the camera coordinate system to the tool coordinate system can be expressed as shown in Eq. (6):

$$^T_C T = {^B_C T}(d_{b\_x}, d_{b\_y}, d_{b\_z}, \delta_{b\_x}, \delta_{b\_y}, \delta_{b\_z}) {^W_B T}(a_i, \alpha_i, d_i, \theta_i) {^T_W T}(d_{t\_x}, d_{t\_y}) \quad (6)$$

## 2.2 Rough identification based on axis measurement

The axis measurement method first needs to obtain the position data of the feature points on the calibration plate, and then fit the space circle to obtain the posture of the axis. In the experiment, the data point is collected by rotating each axis of the seven joint axes of the robot arm separately. The angle is selected to take 10 equidistances in the range of −45° to +45° around the zero-position of the joint to be rotated. The other joints maintain a posture of 0°. After the mechanical arm rotates to the specified angle, it remains stationary for 2 s, and the camera captures the picture to avoid the motion inertia causing the calibration plate to vibrate back and forth around the final position, affecting the quality of the picture collection. Through the library function provided by OpenCV, the pose of the feature points in the calibration plate under each robot arm attitude is determined. A total of 7*10=70 sets of data can be used in the position of 10 spatial poses generated by each axis rotation (X, Y, Z). The space circle is fitted to obtain the pose parameters of each joint axis.

The general equation for a standard space circle is a space curve formed by the intersection of the spherical $\Omega$ in space with the plane $\Pi$ of the passing sphere. The expression is shown in Eq. (7):

$$\Omega:(x-x_0)^2+(y-y_0)^2+(z-z_0)^2=r^2$$
$$\prod:Ax+By+Cz+D=0 \tag{7}$$

Therefore, the equation for determining a space circle needs to solve eight parameters: $A,B,C,D,x_0,y_0,z_0$, and $r$. In this paper, the coordinate transformation method is adopted: a circular coordinate system is established on the plane of the fit, and the spatial point in the measurement coordinate system is converted to the plane where the circle is located by measuring the transformation matrix of the coordinate system to the circular coordinate system. Therefore, to achieve three-dimensional circle fitting, the dimension is reduced to a plane circle, and then the least-squares method is used to fit the plane circle to obtain the geometric parameters of the space circle [13]. The specific algorithm flow is shown in Fig. 2.

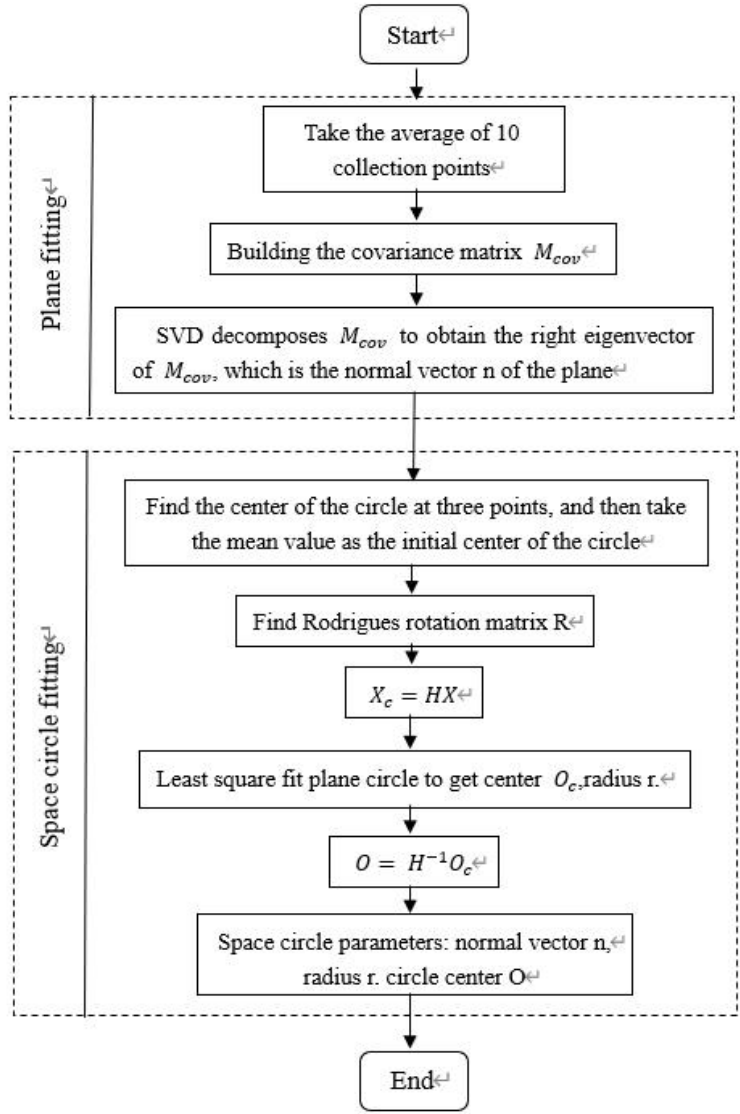

Fig. 2. Fitting of space circles.

The space circle fitting algorithm described above is realized by MATLAB programming, and the rough identification process based on the axis measurement is performed. The plane normal vector, center, and radius of the space circle fitted by the obtained seven joint axes are shown in Table 2, where ($A$, $B$, $C$, $D$) are the parameters of the plane of the space circle, ($x_0, y_0, z_0$) are the coordinates of the center of the fitted space circle in the measurement coordinate system, and $r$ is the radius of the space circle.

Table 2. Parameter table of space circle fitted by each axis.

| Joint | (A,B,C,D) | | | | $(x_0, y_0, z_0,)$ | | | $r$ |
|---|---|---|---|---|---|---|---|---|
| 1 | 0.0040 | 0.9965 | -0.0839 | -53.8488 | -220.2609 | 125.321 | 836.3645 | 164.4061 |
| 2 | -0.9998 | 0.0002 | -0.0218 | -38.5675 | -58.5129 | -731.9035 | 910.7761 | 860.2693 |
| 3 | 0.0021 | 0.9963 | -0.0858 | -52.0788 | -137.3635 | 124.6159 | 836.5622 | 210.1419 |
| 4 | -0.9990 | -0.0003 | -0.0439 | -20.7657 | -59.4907 | -381.7239 | 882.7216 | 509.2803 |
| 5 | 0.0028 | 0.9957 | -0.0926 | -46.6289 | -42.5822 | 124.3529 | 832.3475 | 20.343 |
| 6 | -0.9998 | -0.0037 | -0.0185 | -40.9403 | -56.382 | -86.6378 | 852.2257 | 212.7928 |
| 7 | -0.0111 | 0.9971 | -0.0749 | -62.0452 | -108.7059 | 123.8482 | 836.7923 | 55.4436 |

The fitting result of each axis of the manipulator is shown in Fig. 3. Figure 3(a) shows the spatial circle fitted by the spatial positions of the characteristic points of the calibration plate collected by the first axis and the second axis of the rotating manipulator. Figure 3(b) shows the seven fitted circles and their axes obtained from the spatial positions of the calibration plate feature points acquired by the seven axes of the rotating robot arm.

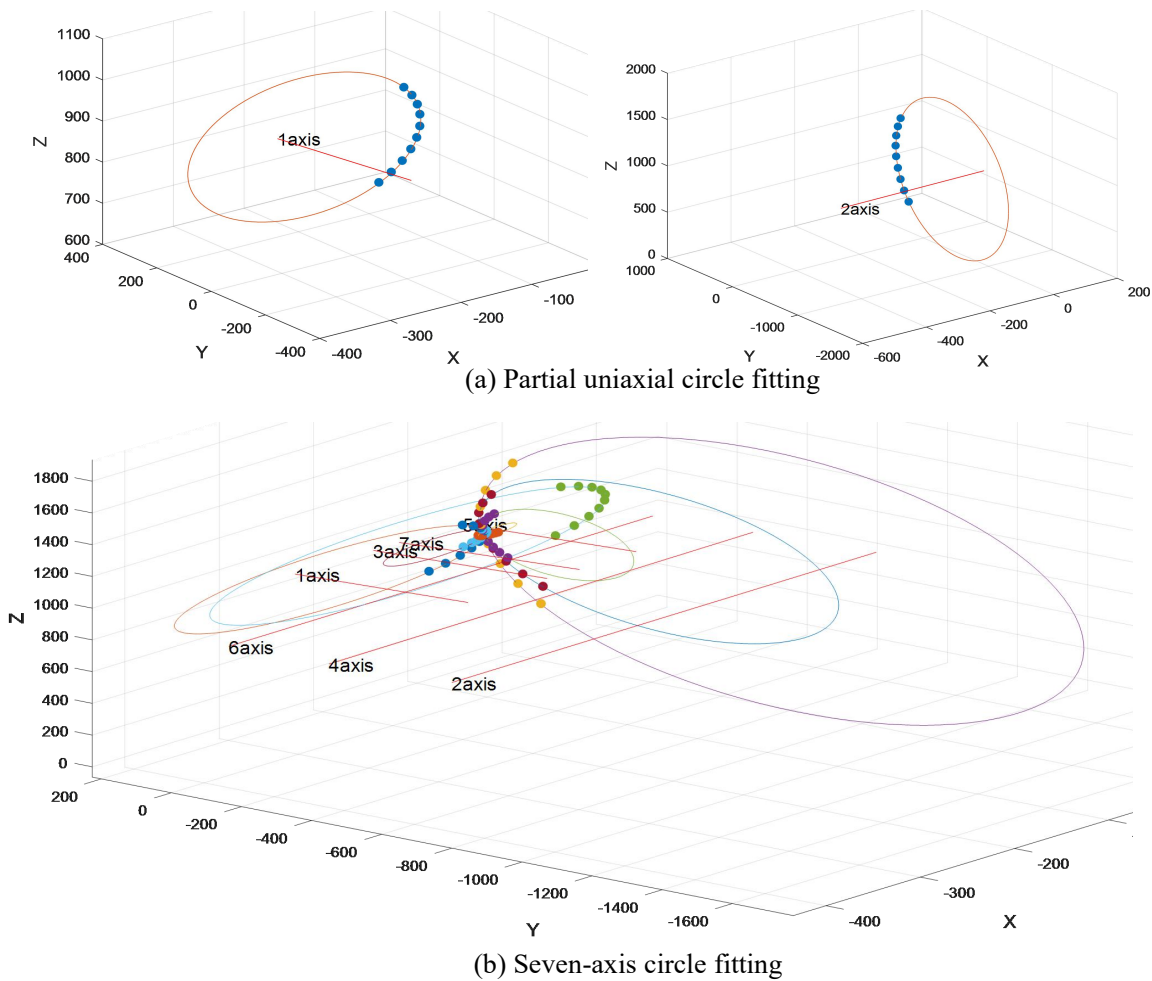

(a) Partial uniaxial circle fitting

(b) Seven-axis circle fitting

Fig. 3. Spatial circle fitting of the data acquired by the robot arm.

After the spatial circle fitting is completed, accuracy analysis is needed. Generally, the common surface distance $d_{cp}$, flatness $m_p$, and roundness $m_c$ are used as evaluation indexes to characterize the fitting precision of the space circle. The cardioid distance represents the distance between the fitted center and the fitted plane. The formula is defined as follows:

$$d_{cp} = |Ax_0+By_0+Cz_0+D|/\sqrt{A^2+B^2+C^2} \tag{8}$$

Flatness represents the mean of the distance between all measured data points and the fitted plane. The formula is defined as follows:

$$m_p = \sqrt{\frac{1}{N}\sum_{i=1}^{N} d_i^2}, \tag{9}$$

where $d_i = |Ax_i+By_i+Cz_i+D|/\sqrt{A^2+B^2+C^2}$ indicates the distance from the $i$th measurement point to the fit plane.

Roundness represents the mean of the distance between all measured data points and the fitted space circle. The formula is defined as follows:

$$m_c = \sqrt{\frac{1}{N}\sum_{i=1}^{N} v_i^2}, \tag{10}$$

where $v_i = \sqrt{(x_i-x_0)^2 + (y_i-y_0)^2 + (z_i-z_0)^2} - r$ indicates the distance from the $i$th measurement point to the fitted space circle. The smaller the above evaluation index value, the higher the accuracy of the representative fitting.

According to the definition formula of the evaluation index of the spatial circle fitting accuracy described above, the accuracy evaluation index for calculating the spatial circle fitting of each axis of the robot arm is shown in Table 3.

Table 3. Evaluation index of fitting accuracy of each arm of the robot arm.

| Joint | $d_{cp}$ | $m_p$ | $m_c$ |
| --- | --- | --- | --- |
| 1 | 1.4211e-14 | 0.017248 | 0.098586 |
| 2 | 7.1054e-15 | 0.11883 | 0.05858 |
| 3 | 0.0 | 0.030578 | 0.18748 |
| 4 | 0.0 | 0.070534 | 0.098626 |
| 5 | 0.0 | 0.027458 | 0.32916 |
| 6 | 0.0 | 0.090841 | 0.10319 |
| 7 | 7.1054e-15 | 0.026229 | 0.23017 |

According to the data analysis in the table, in general, $m_p < m_c$, that is, the plane fitting accuracy is higher than the plane circle fitting accuracy, but in the data of joint axis 2, $m_p > m_c$, and the performance is abnormal. After analysis, because the radius of the fitting circle of joint axis 2 is the largest, at 860 mm, the motion inertia of the calibration plate at the end of the manipulator and the cumulative effect of the geometric error of the kinematic model are large, resulting in the measurement point of the axis. The flatness indexes $m_p$ of joint axes 2, 4, and 6 in the table are greater than those of joint axes 1, 3, 5, and 7, which is the flatness error fitted by joint axes 2, 4, and 6. Because the radius of the circle is larger, the distribution of data points is more dispersed, and the plane fitting error is larger.

After the spatial circle and its axis are fitted, the preliminary D-H parameters can be obtained by solving the length of the common perpendicular line between the axes and the angle of the joint axes, as shown in Table 4. The "-" symbol in the table indicates that this parameter cannot be obtained by the axis measurement method.

Table 4. D-H parameter table obtained by the axis measurement method.

| Joint $i$ | $\alpha_{i-1}(rad)$ | | $a_{i-1}(mm)$ | | $d_i(mm)$ | |
|---|---|---|---|---|---|---|
| | Ideal value | Measurements | Ideal value | Measurements | Ideal value | Measurements |
| 1 | $-\pi/2$ | 1.5728 | 0 | 1.3351 | 85 | - |
| 2 | $-\pi/2$ | 1.5709 | 0 | 1.3064 | 85 | 84.01928 |
| 3 | $\pi/2$ | 1.5695 | 0 | 0.9211 | 350 | 350.9939 |
| 4 | $-\pi/2$ | 1.5698 | 0 | 3.9728 | 100 | 94.80308 |
| 5 | $-\pi/2$ | 1.5755 | 0 | 0.5037 | 300 | 299.0269 |
| 6 | $\pi/2$ | 1.5620 | 0 | 1.3065 | 64 | 62.96186 |
| 7 | 0 | - | 0 | - | 42 | - |

It can be seen from the table that the error of parameter $\alpha_i$ measured by the axis measurement method is within 0.01 rad, the maximum error value of $a_i$ is 3.9728 mm, and the maximum error of $d_i$ is 5.2 mm. Since the axis measured by the axis measurement method is in the zero position, the $\theta$ parameter is 0, and no measurement is made. Due to the different end effectors, $\alpha_7, a_7$, and $d_7$ are different, so these three parameters cannot be obtained by axis measurement.

*2.3 Fine identification based on nonlinear optimization*

The axis measurement method can directly measure the size parameters of the manipulator. It is not necessary to know the initial value in advance, and the versatility is good, but the measurement accuracy is not high, and the radius of rotation and the dispersion of the sample point data have a great influence on the result. Therefore, the D-H parameter measured in the rough identification process based on the axis measurement method is used as the initial value of the fine identification, and then the 36 independent parameters determined in the process of establishing the optimized D-H kinematic model are used as the optimization variables $\Delta$. The actual kinematic model and the optimized motion are obtained. The three-dimensional positions of the calibration plate feature points of each robot arm corresponding to the model are mapped to the pixel coordinate system, the Euclidean distance error of the pixel coordinates is the objective function to be optimized, and a nonlinear function is used to write the objective function. The smallest corresponding optimal variable obtained by fine identification is used in the optimized D-H kinematic model.

The specific process of fine identification is mainly divided into the following steps.

1) Initialization: The rough recognized D-H parameter is taken as the initial value. When the robot arm rotates the picture of the end plate of the arm of each joint axis, the seven joint angles $\theta_{cap}$ in the corresponding posture are recorded. This process is introduced in the camera calibration process. The pixel coordinate $P_{cap}$ of the feature point in the pixel coordinate system in each picture, that is, the pixel coordinate corresponding to the feature point in the actual robot arm position, is determined. The displacement amount of the wrist to the tool coordinate system $^T_W T$ is initially measured using a Vernier caliper. The initial value of the transformation matrix $^B_C T$ of the camera coordinate system to the base coordinate system can be obtained by the axis measurement method: in the experiment, the three-dimensional position of the intersection of axis 1 and axis 2 of the robot arm is used as the displacement of the base coordinate system in the camera coordinates. The amount of displacement under the system, the initial value of the rotation amount, needs to be determined according to the rotation relationship between the camera coordinate system C and the base coordinate system B. As shown in Fig. 4, coordinate system B is first rotated by 90° around the x-axis and then rotated by 90° around the z-axis. In this case, the base coordinate system B is ideally coincident with the camera coordinate system C. The rotation matrix $\boldsymbol{R} = \boldsymbol{Rotx}(90)\boldsymbol{Roty}(-90)$, that is, $\delta_{b\_x} = 90, \delta_{b\_y} = 0, \delta_{b\_z} = 90$, is used as the initial value of the rotation amount in the homogeneous transformation matrix of the two coordinate systems.

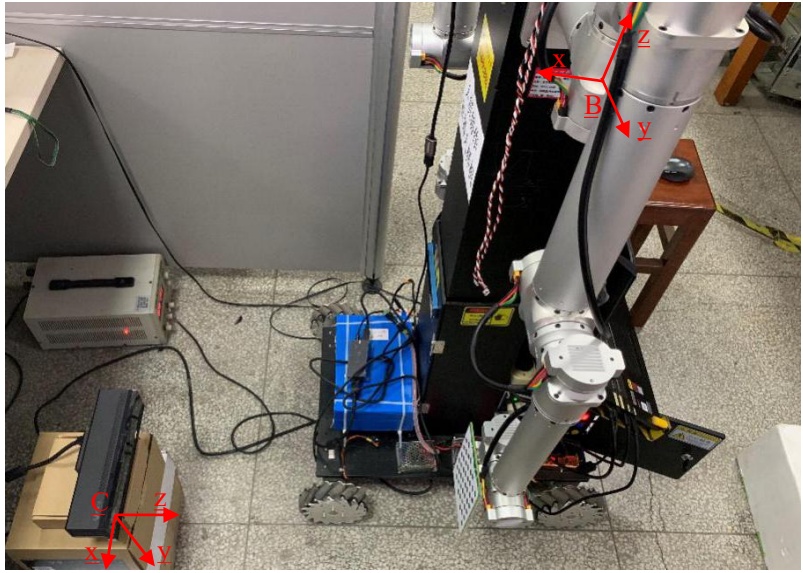

Fig. 4. Schematic diagram of camera coordinate system and robot arm base coordinate system.

2) Update the D-H parameter, the transformation matrix of the camera coordinate system to the base coordinate system ${}^B_C\boldsymbol{T}$, and the transformation matrix of the wrist coordinate system to the tool coordinate system ${}^T_W\boldsymbol{T}$ using the variable Δ (1×36-dimensional matrix) to be optimized. In the parameters, the joint kinematic $\theta_{cap}$ and $DH'$ (updated D-H parameters) are used for the kinematic calculation, and the transformation matrix ${}^7_0\boldsymbol{T}$ of the manipulator base coordinate system to the end of the wrist is obtained. The pixel coordinates $P_{cal}$ of the feature points of the calibration plate in each image corresponding to the optimized kinematic model in the pixel coordinate system, that is, the pixel coordinates corresponding to the feature points in the optimized robot arm pose, can be determined using Eq. (11):

$$P_{cal} = \boldsymbol{K}\,{}^B_C\boldsymbol{T}\,{}^7_0\boldsymbol{T}\,{}^T_W\boldsymbol{T} \tag{11}$$

where **K** is the extended camera internal reference matrix, and ${}^B_C\boldsymbol{T}, {}^T_W\boldsymbol{T}, {}^7_0\boldsymbol{T}$ is the coordinate system transformation matrix after optimizing the D-H kinematic model. Then the objective function is taken as the minimum value of the sum of the modulus of the difference between the coordinates $P_{cap}$ of the calibration plate feature point in the pixel coordinate system and the coordinates $P_{cal}$ calculated using the optimized D-H kinematic model in the actual pose of the robot arm, shown in Eq. (12):

$$\min \sum_{i=1}^{N} \|P_{cap} - P_{cal}\| \tag{12}$$

Compared with traditional high-precision three-dimensional measuring instruments, the monocular camera-based visual calibration technology has more uncertainties and systematic errors. In order to avoid excessive error, the three-dimensional position of the calibration plate feature points determined by the three-dimensional position of the actual calibration plate feature point and the optimized D-H parameter is mapped to the pixel by the calibrated camera internal reference matrix. The distance calculation is performed in the coordinate system. Compared with the traditional method, the position deviation of the end effectors of the two ends is directly compared with the target function, which can avoid the systematic error caused by the camera parameters in the visual calibration. The specific implementation process is shown in Fig. 5.

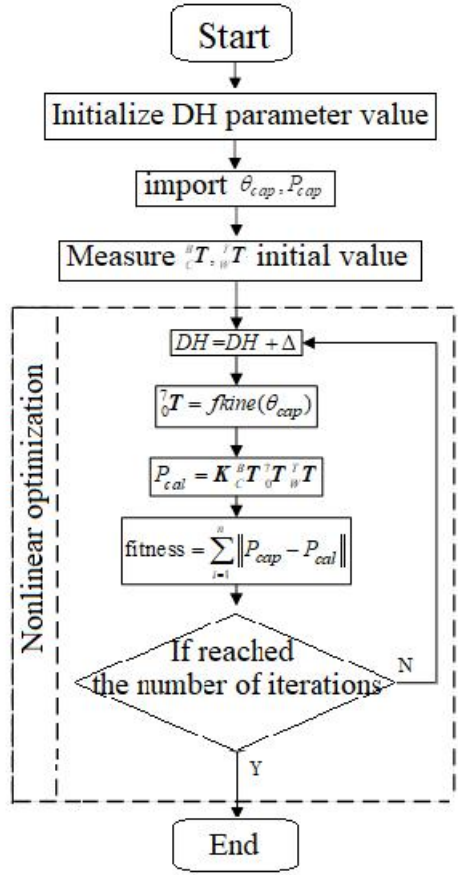

Fig. 5. The fine identification flowchart.

Because the number of variables to be optimized is too high, the selection of nonlinear optimization algorithm becomes the key to calibration accuracy. The objective function of this paper is the coordinate difference of the calibration plate feature points in the pixel coordinate system corresponding to the two kinematic models. It is an unconstrained and non-convex nonlinear least-squares problem. For the occasions where more variables must be optimized, the correlation is not strong, and the function is not convex, traditional algorithm such as the genetic algorithm and the swarm intelligence optimization algorithm are not effective, and the optimization speed is slow; therefore, this paper uses the quasi-Newton method for optimization as it can converge to the final result more quickly.

The D-H parameter in Table 4 is taken as the initial value of the D-H parameter of the fine identification process. According to the algorithm flowchart shown in Fig. 5, MATLAB is used to write the nonlinear optimization objective function, and the optimization toolbox of MATLAB is used to optimize the objective function for the unconstrained nonlinear optimization solver *fminunc*. The *fminunc* solver can be used internally. Two algorithms, quasi-Newton method and confidence domain method, are implemented. The confidence domain method needs to deduct the objective function, and its derivative is one of the output parameters of the function. However, the objective function of this paper involves more variables, and the derivation process is more complicated. Therefore, the quasi-Newton method is used to realize nonlinear optimization. The optimized iterative process is shown in Fig. 6. It can be seen that the optimized objective function value is 0.995811; that is, the average deviation of the calibration plate feature points at the end of the manipulator after calibration from the actual values in the pixel coordinate system is 1 pixel. The corresponding optimization variable values when the objective function is minimized are shown in Tables 5 and 6.

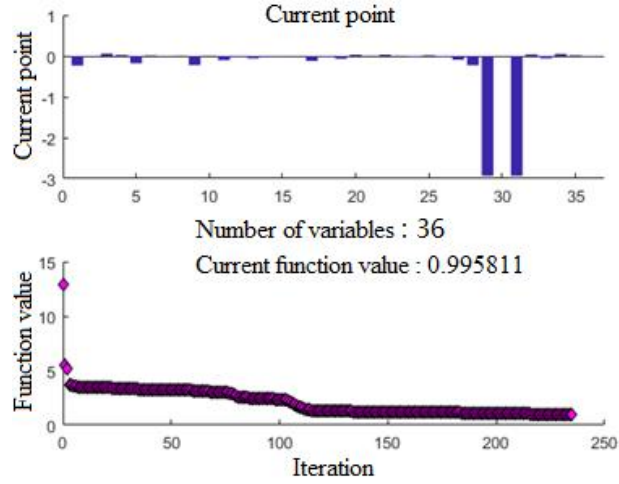

Fig. 6. Precision identification nonlinear optimization iteration graph.

Table 5. Optimizing D-H kinematic parameter offset.

| Joint $i$ | $\Delta\alpha_{i-1}(rad)$ | $\Delta a_{i-1}(mm)$ | $\Delta d_i(mm)$ | $\Delta\theta_i(°)$ |
|---|---|---|---|---|
| 1 | -0.2355 | -0.0071 | 0.0574 | 0.0221 |
| 2 | -0.1798 | 0.0096 | -0.0082 | 0.0026 |
| 3 | -0.2247 | 0.0034 | -0.1014 | -0.0230 |
| 4 | -0.0551 | -0.0130 | -0.0153 | -0.0087 |
| 5 | -0.1206 | -0.0211 | -0.0694 | 0.0284 |
| 6 | 0.0056 | 0.0303 | 0.0044 | -0.0275 |
| 7 | 0.0164 | -0.0287 | -0.0905 | -0.2285 |

Table 6. Conversion deviation between coordinate systems.

| | | | |
|---|---|---|---|
| Displacement deviation from the base coordinate system to the camera coordinate system (mm) | $d_{b\_x}$ | $d_{b\_y}$ | $d_{b\_z}$ |
| | 0.0391 | -0.0576 | 0.0483 |
| Rotational deviation of the base coordinate system to the camera coordinate system (°) | $\delta_{b\_x}$ | $\delta_{b\_y}$ | $\delta_{b\_z}$ |
| | -2.9305 | -2.9305 | -0.0106 |
| Displacement deviation from wrist to tool coordinate system (mm) | $d_{t\_x}$ | | $d_{t\_y}$ |
| | 0.0164 | | -0.0143 |

As can be seen from Fig. 6 and Table 6, the main deviation is the large deviation (−2.9305°) of the rotation amount of the robot arm base coordinate system with respect to the camera coordinate system. This is because the camera cannot ensure that the XOY plane of the coordinate system C is completely parallel with the plane of the robot arm zero position.

## 3. Experimental System Design

The calibration experiment system of this paper is shown in Fig. 7. The camera is an RGB monocular camera in Kinect V2. Compared with traditional industrial cameras, Kinect V2 is more convenient to use and maintain.

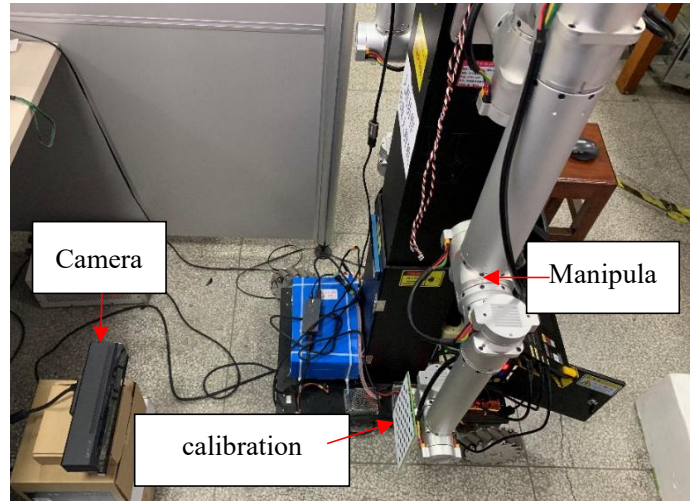

Fig. 7. Calibration experiment system.

The calibration plate is fixed to the end of the seventh joint of the mechanical arm by a connecting member, the mechanical arm is mounted on the lifting column of the moving chassis, and the moving chassis and the lifting column are kept stationary during the calibration experiment.

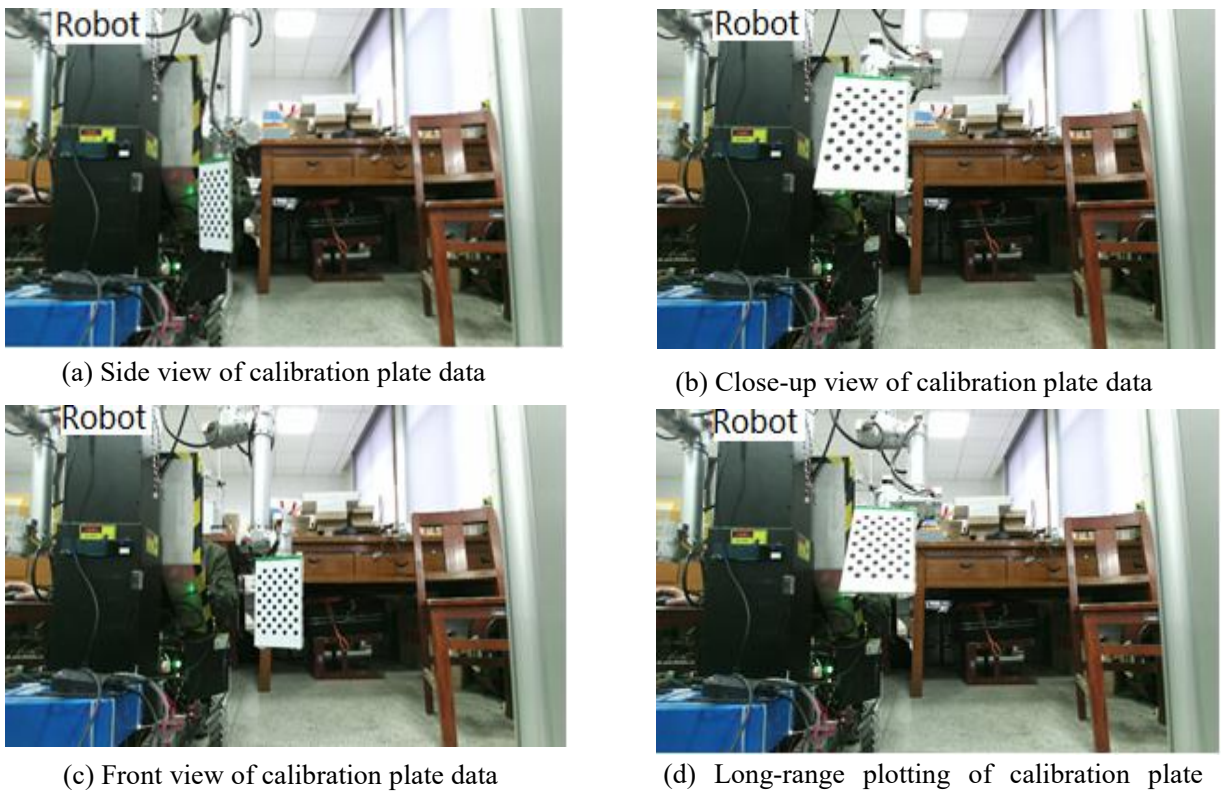

(a) Side view of calibration plate data

(b) Close-up view of calibration plate data

(c) Front view of calibration plate data

(d) Long-range plotting of calibration plate

Fig. 8. Collection of the calibration plate data map.

The process of collecting the calibration plate data is as follows. The initial state of the arm is the zero position. Starting from the first joint in its zero position (to ensure that the picture captured by the camera can detect the corner point), the joint is rotated 10 times between −45° and +45°, each time at a fixed angle, stopping for 2 s after each rotation, and using the camera to shoot the calibration plate at this time; when one joint has been rotated 10 times, it is returned to its zero position. Then the next joint rotation starts from −45°. Once all seven joints have been rotated 10 times and a total of 70 pictures have been collected, imaging is stopped. Some of the captured images are shown in Fig. 8. Since the calibration plate fixed at the end of the robot arm occupies a small range in the field of view of the camera, in order to obtain a more accurate parameter of the internal reference matrix, it is necessary to collect additional images with different poses and a larger field of view. Hence, additional 12 calibrations are collected in this article. The board is not fixed at the end of the arm, and the picture with a larger field of view is used for the internal reference calibration.

## 4. Experimental Verification

After the precise identification process, a set of optimized variables Δ can be obtained, including parameters in D-H parameters, $^{B}_{C}T$ and $^{T}_{W}T$, and the results of the rough identification are compensated using Δ, and the classic D-H is compared. The differences between the actual measured calibration plate feature point pose and the calibration plate feature point poses determined by the classic D-H kinematic model and the optimized D-H kinematic model are calculated to verify that the monocular camera-based positive kinematic calibration technique improves absolute positioning.

The main purpose of this experiment is to calibrate the D-H parameters of kinematics. Therefore, in order to avoid the comparison between the transformation matrix $^{B}_{C}T$ and the wrist coordinate system of the base coordinate system and the camera coordinate system to the tool coordinate system $^{T}_{W}T$, the calibration results are compared. The effect is obtained by compensating $^{B}_{C}T$ and $^{T}_{W}T$ rough values with the parameters in Table 6 to obtain $^{B}_{C}T'$ and $^{T}_{W}T'$. The subsequent comparison experiments use the compensated $^{B}_{C}T'$ and $^{T}_{W}T'$ to solve the position of the feature points of the end plate of the arm.

The D-H parameter offset in Table 5 after optimization is used to compensate the D-H parameter obtained by the rough identification to obtain $DH'$, and the theoretical D-H parameter and the rough identification (axis measurement method) are used to obtain the D-H parameter (i.e., D-H parameter after compensation). The former D-H parameter, the parameter obtained by the fine identification $DH'$ (i.e., D-H parameter before compensation), is used to establish the kinematic model of the manipulator, and the actual joint angle data $\theta_{cap}$ are used to solve the positive kinematics. Using Eq. (11), the Euclidean distance is calculated between the pixel coordinate $P_{cal}$ of the calibration plate feature point and the pixel coordinate $P_{cap}$ of the calibration plate feature point corresponding to the actual mechanical arm configuration under the different D-H parameters.

## 5. Experimental Results and Analysis

The smaller the Euclidean distance of the pixel coordinates of the calibration plate feature points corresponding to the position of each mechanical arm under different D-H parameters, the closer the actual value is, that is, the higher the calibration accuracy, as shown in Fig. 9.

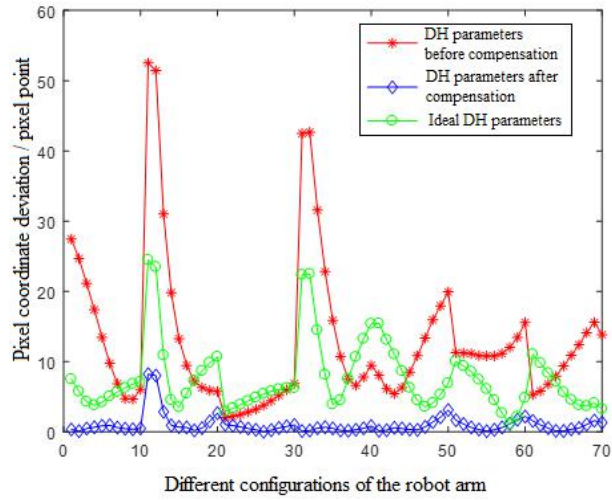

Fig. 9. Comparison of pixel deviations between different D-H parameters and actual values.

It can be seen that the pixel coordinates calculated according to the classic D-H parameters deviate from the actual pixel coordinates, and the average deviation is calculated to be 7.9 pixels. The average deviation of the pixel coordinates calculated by the D-H parameters before compensation is 12.98 pixels, and the average deviation of the pixel coordinates calculated by the D-H parameters after compensation (i.e., the optimized D-H kinematic model) is 0.99 pixels. Therefore, pixel deviation is reduced by nearly 7 pixels when using the optimized D-H parameters after compensation compared to the classic D-H parameters. Thus, the absolute positioning accuracy of the robot arm is significantly improved.

Then the actual measurement position of the feature points on the calibration plate obtained by the OpenCV library during the calibration of the internal reference of the camera, the three-dimensional position of the feature points of the calibration plate according to the D-H parameters before compensation, and the positive motion according to the compensated D-H parameters are compared. The three-dimensional positions of the characteristic points of the calibration plate as well as the three-dimensional positions of the characteristic points of the calibration plate according to the classic D-H parameters are calculated and drawn using MATLAB, as shown in Fig. 10. It can be seen that the actual measurement position is basically consistent with the three-dimensional position data points measured by the compensated D-H parameters, and the error between the three-dimensional position and the actual measurement position measured according to the classic D-H parameters and the D-H parameters before compensation is large. Therefore, the proposed visual calibration method can effectively improve the absolute positioning accuracy of the end of the arm.

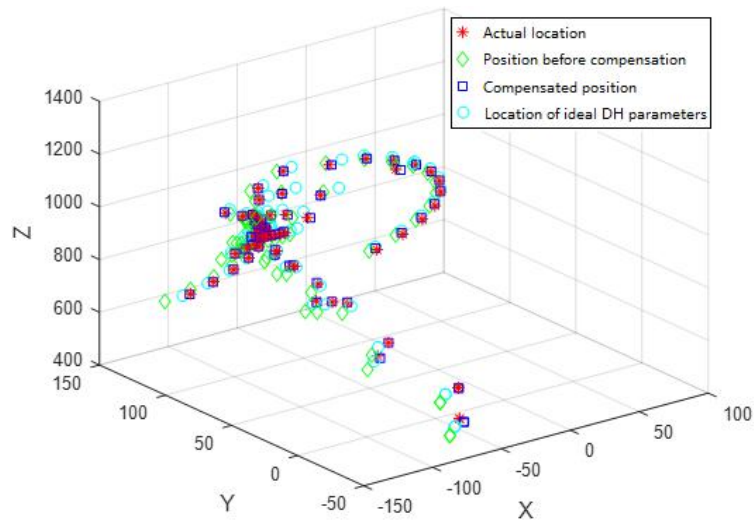

Fig. 10. Comparison of position and actual position under different D-H parameters.

## 6. Conclusion

This paper proposes a visual kinematics calibration method for manipulator based on nonlinear optimization. Using the visual information obtained by the monocular camera, the three-dimensional position of the feature points of the calibration plate under each robot arm attitude corresponding to the actual kinematic model and the classic D-H kinematic model is mapped to the pixel coordinate system, and the sum of the Euclidean distance errors of the two pixel coordinates is used as the objective function to be optimized. At the same time, the actual deviation of the D-H parameter of the robotic arm is obtained through two steps of rough identification based on the axis measurement method and precise identification based on nonlinear optimization, and then the D-H parameter of the robotic arm is compensated to obtain the optimized D-H parameter. This method can effectively avoid system errors caused by camera parameters in visual calibration and improve the absolute positioning accuracy of the end of the robotic arm. At the same time, the method has good economics and universality in the calibration of robotic kinematics.

**Gang Peng**, received the doctoral degree from the Department of control science and engineering of Huazhong University of Science and Technology (HUST) in 2002. Currently, he is an associate professor in the Department of Automatic Control, School of Artificial Intelligence and Automation, HUST. He is also a senior member of the China Embedded System Industry Alliance and the China Software Industry Embedded System Association, a senior member of the Chinese Electronics Association, and a member of the Intelligent Robot Professional Committee of Chinese Association for Artificial Intelligence. His research interests include intelligent robots, machine vision, multi-sensor fusion, machine learning and artificial intelligence.

**Zhihao Wang**, received bachelor degree in school of automation from Wuhan University of Technology in 2017. He received master degree at the Department of Automatic Control, School of Artificial Intelligence and Automation, Huazhong University of Science and Technology. His research interests include machine vision and intelligent robots.

**Jin Yang,** received bachelor degree in school of automation from Tianjin Polytechnic University in 2019. He is currently a graduate student at the Department of Automatic Control, School of Artificial Intelligence and Automation, Huazhong University of Science and Technology. His research interests include intelligent robots, deep learning and artificial intelligence.

**Xinde Li**, professor and doctoral tutor, graduated from the Control Department of Huazhong University of Science and Technology in June 2007, worked for the School of Automation of Southeast University in December of the same year. From January 2012 to January 2013 as a national public visiting scholar at Georgia Tech Visit and exchange for one year. He was selected as an IEEE Senior member in 2016. From January 2016 to the end of August 2016, he worked as a Research Fellow in the ECE Department of the National University of Singapore. His research interests include intelligent robots, machine vision perception, machine learning, human-computer interaction, intelligent information fusion and artificial intelligence.